\documentclass{article}
\usepackage[utf8]{inputenc}
\usepackage{nodalida2015}
\usepackage{mathptmx}
\usepackage{url}
\usepackage{times}
\usepackage{latexsym}
\usepackage{graphicx}
\usepackage{comment}
\usepackage{booktabs}
\usepackage{amsmath}
\usepackage{hyperref}
\usepackage{caption}
\usepackage{subcaption}

\usepackage{natbib}
\bibpunct{(}{)}{,}{a}{ }{;}

\usepackage{sparklines}
\definecolor{sparkrectanglecolor}{rgb}{0.8,0.95,0.8}
\definecolor{sparkspikecolor}{rgb}{1,0,0}
\title{Inferring the location of authors from words in their texts}

\author{Max Berggren \& Jussi Karlgren \\
  Gavagai, 
  Stockholm \\
  {\tt \{max, jussi\}@gavagai.se} \\ \And
  Robert \"Ostling \& Mikael Parkvall \\ Dept of Linguistics,
Stockholm University \\ {\tt \{robert, parkvall\}@ling.su.se}}

\begin{document}

\maketitle
\begin{abstract}
For the purposes of computational dialectology or other geographically bound text analysis tasks, texts must be annotated with their or their authors' location. Many texts are locatable but most have no explicit annotation of place. This paper describes a series of experiments to determine how positionally annotated microblog posts can be used to learn location indicating words which then can be used to locate blog texts and their authors. A Gaussian distribution is used to model the locational qualities of words.  We introduce the notion of placeness to describe how locational words are.  

We find that modelling word distributions to account for {\em several locations} and thus several Gaussian distributions per word, defining a filter which picks out words with high placeness based on their {\em local distributional context}, and aggregating locational information in a {\em centroid} for each text gives the most useful results. The results are applied to data in the Swedish language. 
\end{abstract}

\section{Text and Geographical Position}
Authors write texts in a location, about something in a location (or about the location itself), reside and conduct their business in various locations, and have a background in some location. Some texts are personal, anchored in the here and now, where others are general and not necessarily bound to any context. Texts written by authors reflect the above facts explicitly or implicitly, through explicit author intention or incidentally. 
When a text is locational, it may be so because the author mentions some location or because the author is contextually bound to some location. In both cases, the text may or may not have explicit mentions of the context of the author or mention other locations in the text.

For some applications, inferring the location of a text or its author automatically is of interest. We present in this paper how establishing the location of a text can be done by the locational qualities of the terminology used by its author. Here, we investigate the utility of doing so for two distinct use cases.

Firstly, for detecting regional language usage for the purposes of real-time dialectology. The issue here is to find differences in term usage across locations and to investigate whether terminological variation differs across regions. In this case, the ultimate objective is to collect sizeable text collections from various regions of a linguistic area to establish if a certain term or turn of phrase is used more or less frequently in some specific region. The task is then to establish where the author of a text originally is from. This has hitherto been investigated by manual inspection of text collections. \citep[e.g.]{parkvall2012kartor}

Secondly, for monitoring public opinion of e.g. brands, political issues, or other topic of interest. In this case the ultimate objective is to find whether there is a regional variation for the occurrence of opinionated mentions for the topic or topical target under consideration. The task is then to establish the location where a given text is written, or, alternatively, what location the text refers to. 

In both cases, the system is presented with a body of text with the task of assigning a likely location to it. In the former task, typically the body of text is larger and noisier (since authors may refer to other locations than their immediate context); in the second task, the text may be short and have little evidence to work from. Both tasks, that of identifying the location of an author, or that of a text, have been addressed by recent experiments with various points of departure: knowledge-based, making use of recorded points of interest in a location, modelling the geographic distribution of topics, or using social network analysis to find additional information about the author. 

This set of experiments focuses on the text itself and on using distributional semantics to refine the set of terms used for locating a text. 

\section{Location and words as evidence of locations}
Most words contribute little or not at all to positioning text. Some words are dead giveaways: an author may mention a specific location in the text. Frequently, but not always, this is reasonable evidence of position. Some words are less patently locational, but contribute incidentally, such as the name of some establishment or some characteristic feature of a location.

\begin{figure}[ht]
\centering
    \includegraphics[width=0.45\textwidth]{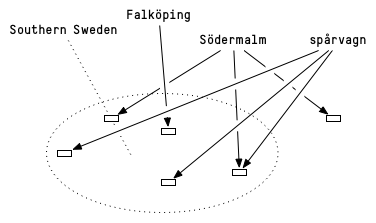}
    \caption{Some terms are polylocational}
    \label{termtypes}
\end{figure}

Some locational terms are polysemous; some inspecific; some are vague. As indicated in Figure~\ref{termtypes}, the term {\it Falköping} unambiguously indicates a town in {\it Southern Sweden}, which in turn is a vague term without a clear and well defined border to other bits of Sweden. The term {\it Södermalm} is polysemous  and refers to a section of town in several Swedish towns; the term {\it spårvagn} (``tram'') is indicative of one of several Swedish towns with tram lines. We call both of these latter types of term {\em polylocational} and allow them to contribute to numerous places simultaneously. 

Other words contribute variously to location of a text. Some words are less patently locational than named places, but contribute incidentally, such as the name of some establishment, some characteristic feature of a location, some event which takes place in some location, or some other topic the discussion of which is more typical in one location than in another. We will estimate the {\em placeness} of words in these experiments.

\section{Mapping from a continuous to a discrete representation}

We, as has been done in previous experiments, collect the geographic distribution of word usage through collecting microblog posts, some of which have longitude and latitude, from Twitter. Posts with location information are distributed over a map in what amounts to a continuous representation. The words from posts can be collected and associated with the positions they have been observed in. 

First experiments which use similar training data to ours have typically assigned the posts and thus the words they occur in directly to some representation of locations - a word which occurs in tweets at $[N59.35,E18.11]$ and $[N59.31,E18.05]$ will have both observations recorded to be in the same city \citep{cheng2010you,mahmud2012tweet}. An alternative and later approach by e.g. \citet{priedhorsky2014inferring} is to aggregate all observations of a word over a map and assign a named location to the distribution, rather than to each observation, deferring the labeling to a point in the analysis where more understanding of the term distribution is known. 

Another approach is to model {\em topics} as inferred from vocabulary usage in text across their geographical distribution, and then, for each text, to assess the topic and thus its attendant location visavi the topic model most likely to have generated the text in question \citep{eisenstein2010latent,yin2011geographical,kinsella2011m,hong2012discovering}. We have found that topic models as implemented are computationally demanding, do not add accuracy to prediction, and have little explanatory value to aid the understanding of localised language use. 

In these experiments we will compare using a list of known places with a model where we aggregate the locational information provided by words  (and potentially other linguistic items such as constructions) trained on longitude and latitude either by letting the words vote for place or by averaging the information on a word-by-word basis. The latter model defers the mapping to place until some analysis has been performed; the former assigns place to the words earlier in the process. 

\section{Test Data}

\begin{figure}[h]
\centering
    \includegraphics[width=0.15\textwidth]{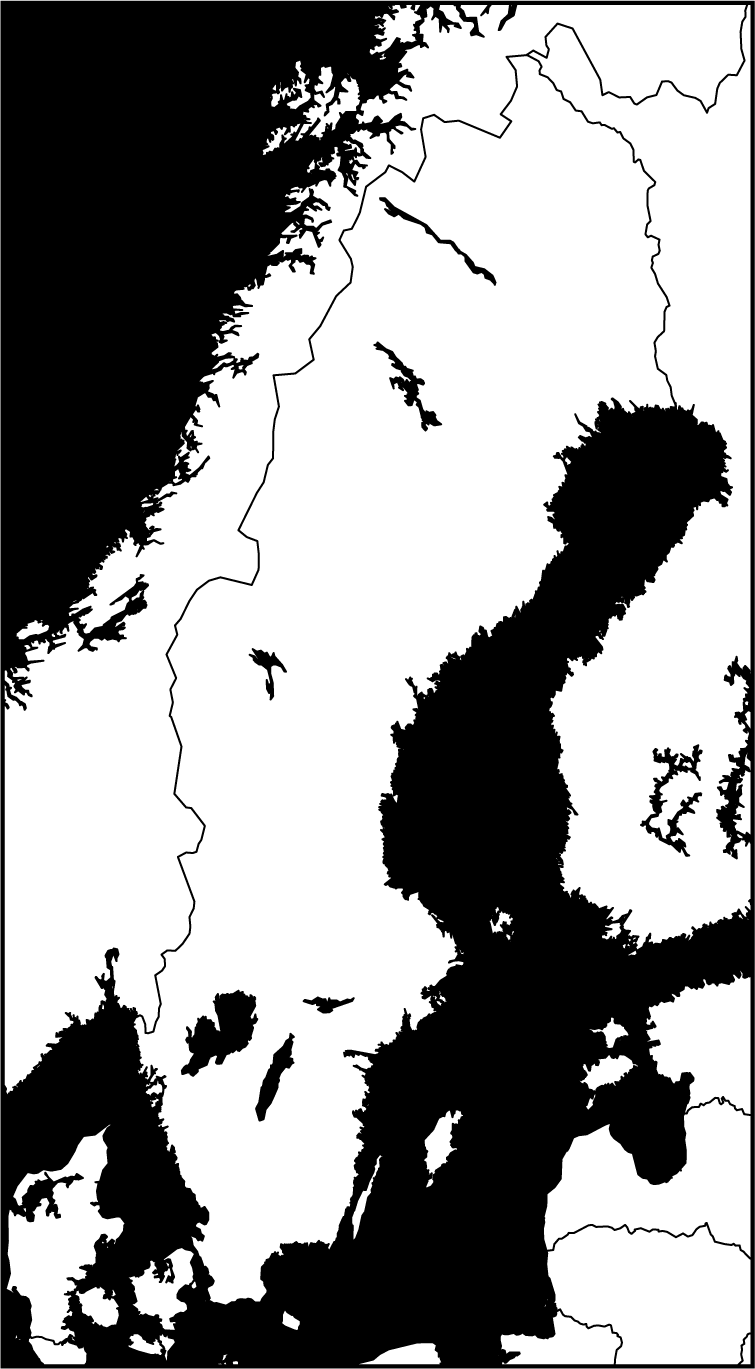}
    \caption{Map of Sweden}
    \label{swedenmap}
\end{figure}

These experiments have focused on Swedish-language material and on Swedish locations. Most Swedish-speakers live in Sweden; Swedish is mainly written and spoken in Sweden and in Finland. Sweden is a roughly rectangular country of about $450~000~km^2$ as shown in Figure~\ref{swedenmap}. 
Sweden has since 1634 been organised into 22 counties or {\em län} of between $3~000~km^2$ and $100~000~km^2$. The median size of a county is $10~545~km^2$ which would, assuming quadratic counties, give a side of $100~km$ for a typical county.

We measure accuracy of textual location using the {\em Haversine distance}, the great-circle distance between two points on a sphere. We report averages, both mean and median, as well as percentage of texts we have located within $100~km$ from their known position. 

Our test data set is composed of social media texts. Firstly, 18 GB of blog text from major Swedish blog and forum sites, with self-reported location by author - variously, home town, municipality, village, or county. The texts are mainly personal texts with authors of all ages but with a preponderance of pre-teens to young adults. The data are from 2001 and onward, with more data from the latest years. The data are concatenated into one document per blog, totalling to $154~062$ documents from unique sources. Somewhat more than a third, 35\%, have more than 10k characters. 

Secondly, 37 GB of blog text without any explicit indication of location. A target task for these experiments is to enrich these 37 GB of non-located data with predicted location, in order to address data sparsity for unusual dialectal linguistic items. 

\section{Baseline: the {\sc Gazetteer} model}

For a list of known places we used a list\footnote{http://en.wikipedia.org/wiki/List\_of\_urban\_areas\_in\_Sweden \\ One named location (``När'') was removed from the list since it is homographic to the adverbials corrresponding to the English {\it near} and {\it when}, causing a disproportionate amount of noise.} of 1~956 Swedish cities and 2~920 towns and villages as defined by Statistics Sweden\footnote{\textit{A locality consists of a group of buildings normally not more than 200 metres apart from each other, and must fulfil a minimum criterion of having at least 200 inhabitants. Delimitation of localities is made by Statistics Sweden every five years.} [http://www.scb.se]} in 2010. 

As the most obvious baseline, we identify all tokens found in the gazetteer. Each such token is converted to a position through the Geoencoding API offered by Google\footnote{\href{https://developers.google.com/maps/documentation/geocoding/}{https://developers.google.com/.../geocoding/}}. The position with largest observed frequency of occurrence in the text is assumed to be the position of the text. Other approaches have taken this as a useful approach for identifying features such as Places of Interest mentioned in texts \citep{li2014effective}. We call this approach the {\sc Gazetteer} approach. 

\section{Training Data}\label{trainingdata}

As a basis for learning how words were used we used geotagged microblog data from Twitter. About 2\% of Swedish Twitter posts have latitude and longitude explicitly given,\footnote{Determined by listening to Twitter's streaming API for about a day.} typically those that have been posted from a mobile phone. We gathered data from Twitter's streaming API\footnote{The ``garden hose'': https://dev.twitter.com/streaming/public \ } during the months of May to August of 2014, saving posts with latitude and longitude and with Sweden explicitly given as point of origin. This gave us 4~429~516 posts of about 630 MB. 

\section{Polylocational Gaussian Mixture Models}

Given a set of geographically located texts, we record for each linguistic item -- meaning word, in these experiments -- the locations from the metadata of every text it occurs in. This gives each word a mapped geographic distribution of latitude-longitude pairs. We model these observed distributions using Gaussian 2-D functions, as defined by \citet{priedhorsky2014inferring}. A 2-D Gaussian function will assume a peak at some position and allow for a graceful inclusion of hits at nearby positions into the model in a bell-like distribution. 

\begin{figure}[ht]
\centering
    \includegraphics[width=0.45\textwidth]{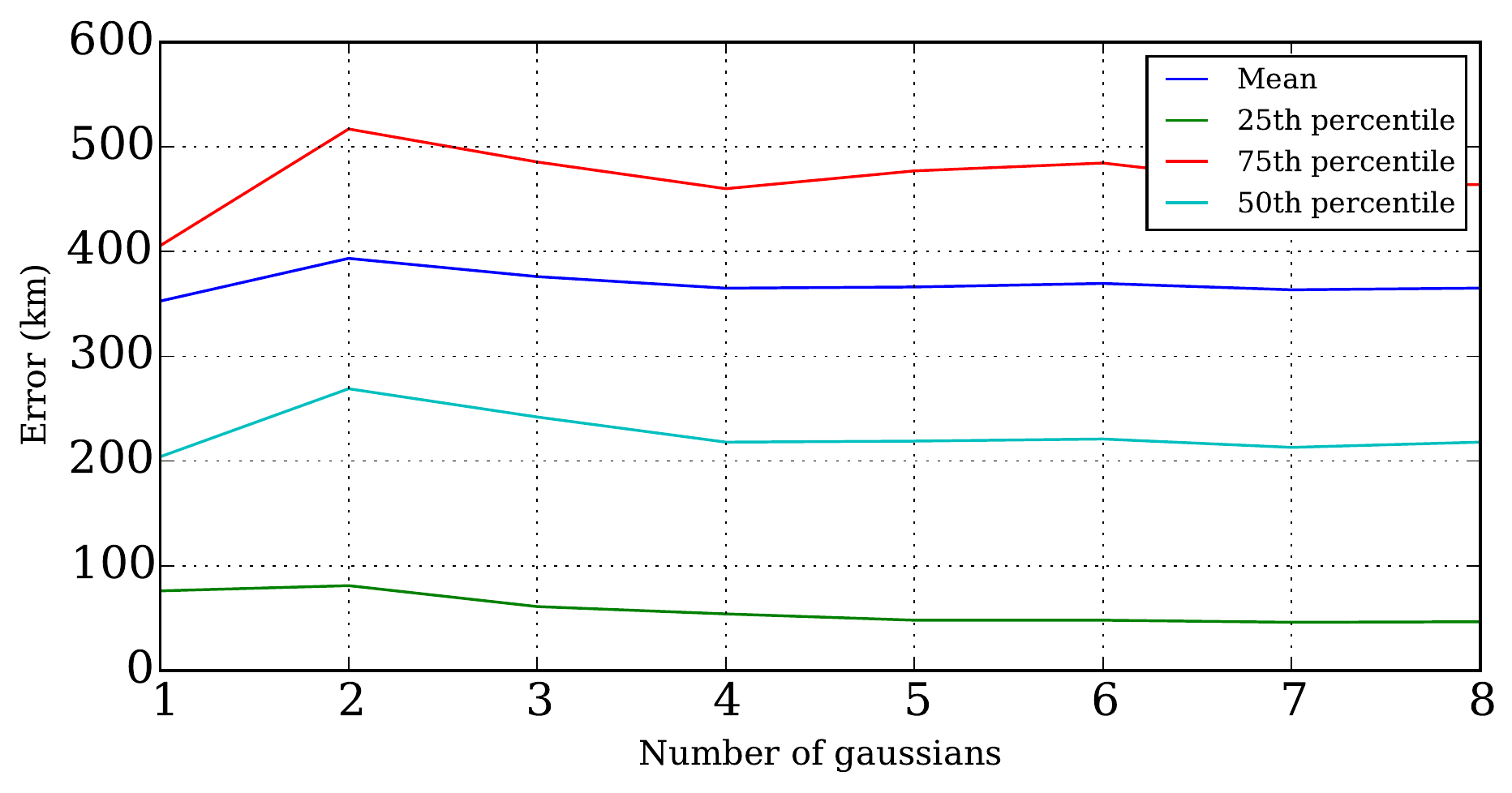}
    \caption{Effect of allowing polylocational representations}
    \label{polygauss}
\end{figure}

In contrast to the original definition and and other similar following approaches, we want to be able to handle polylocational words. After testing various models on a subset of our data we find that fitting more than one Gaussian function---in effect, assuming that locationally interesting words refer to several locations--yields better results than fitting all locational data into one distribution. After some initial parameter exploration as shown in Figure~\ref{polygauss}, we settle on three Gaussian functions as a reasonable model: words with more than three distributional peaks are likely to be of less utility for locating texts. We consequently fit each word with three Gaussian functions to allow a word to contribute to many locations for the texts it is observed in.

\section{The notion of placeness}
\label{placeness}

In keeping with previous research on geolocational terms such as \citet{han2014text}, we rank candidate words for their locational specificity. From the Gaussian Mixture Model representation, we take the log probability $\rho$ in the mean of the Gaussian and transform it into a {\it placeness} score by $ p = e^{\frac{100}{-\rho}}$. This is done for every word, for all three Gaussians. The score is then used to rank words for locational utility. 

\begin{table}[h]
\centering
\begin{tabular}{cc|c|c|c}
 & & \multicolumn{3}{c}{Gaussian} \\
 & & 1st & 2nd & 3d \\
\hline
Falköping & & 58 & 9 & 9  \\
Stockholm & & 37 & 10 & 10 \\
 spårvagn & {\it ``tram''} & 36 & 18 & 15  \\
 och & {\it ``and''} & 16 & 15 & 9 
\end{tabular}
\caption{Example words and their log placeness}
\label{samplewordplaceness}
\end{table}

Table~\ref{samplewordplaceness} shows the placeness of the three Gaussians for some sample words. The two sample named locations have high placeness for their first Gaussians, indicating that they have locational utility. ``Stockholm'', the capital city, which is frequently mentioned in conversations elsewhere has less placeness than has ``Falköping'', a smaller city.  The word ``tram'' has lower placeness than the two cities, and the word ``and'' with a log placeness score of 16 can not be considered locational at all. Inspecting the resulting list as given in Table~\ref{samplehighplaceness} which shows some examples from the top of the list, we find that words with high placeness frequently are non-gazetteer locations (``Slottsskogen''), user names, hash tags -- frequently referring to events (``\#lundakarneval''), and other local terms, most typically street names (``Holgersgatan''), spelling variants (``Ståckhålm''), or public establishments.

The performance of the predictive models introduced below can be improved by excluding words with low placeness from the centroid. This exclusion threshold is referred to as $T$ below.

\begin{table}[h]
{\small
\centering
\begin{tabular}{c|c|c}
 known places & hash tags & other \\
\hline
hogstorp & \#lundakarneval  & holgersgatan \\
nyhammar & \#bishopsarms & margrete\-gärdeparken \\
sjuntorp & \#gothenburg & uddevallahus \\
tyringe & \#westpride14 & kampenhof \\
slottsskogen & \#swedenlove1dday & ståckhålm \\
storvik & \#sverigemotet & gullmarsplan \\
charlottenberg  & \#sthlmtech & tvärbanan \\
\hline
\end{tabular}
}
\caption{Example words with high placeness}
\label{samplehighplaceness}
\end{table}

\begin{figure}[h]
\centering
\footnotesize
\begin{subfigure}{0.35\textwidth}
\includegraphics[width=\textwidth]{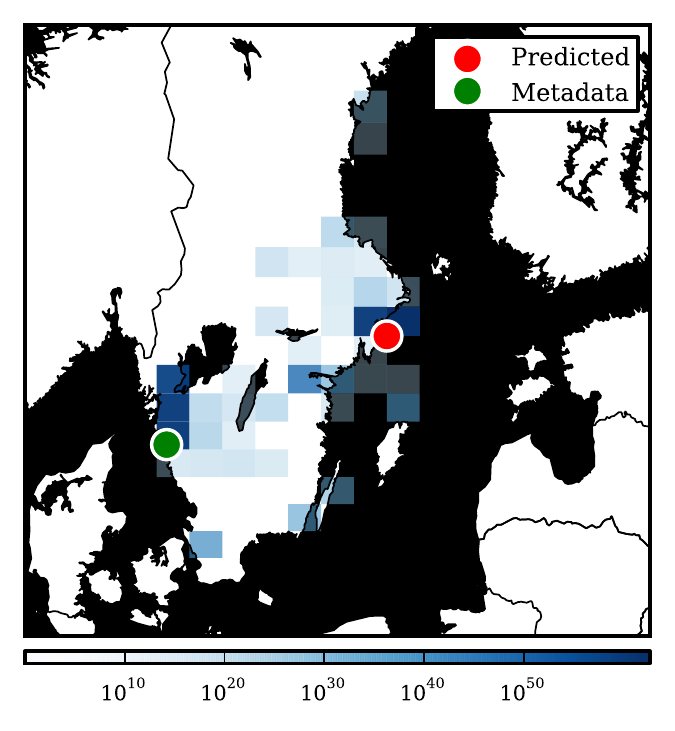}
\caption{All words of a text contribute to the predicted location  \includegraphics[width=0.02\textwidth]{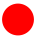}.}
\end{subfigure} \quad \begin{subfigure}{0.35\textwidth}
\includegraphics[width=\textwidth]{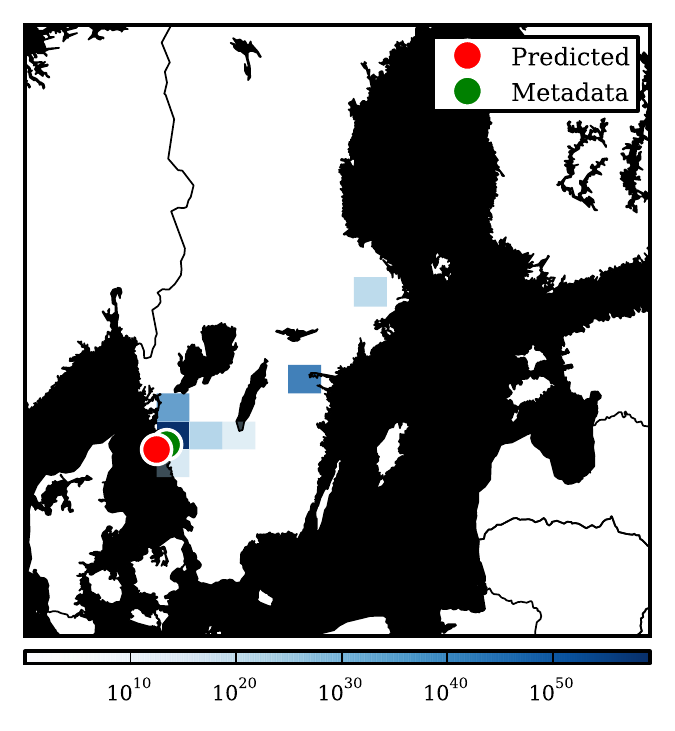}
\caption{Only words filtered through the distributional model contribute votes to yield a prediction \includegraphics[width=0.02\textwidth]{reddot.png} very close to the correct position \includegraphics[width=0.02\textwidth]{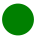}.}
\end{subfigure}
\caption{Comparison illustrating the grid and showing how the grammar transforms the result.} \label{gridexample}
\end{figure}

\begin{figure*}[t]
\centering
\footnotesize
\begin{subfigure}{0.75\textwidth}
    \includegraphics[width=\textwidth]{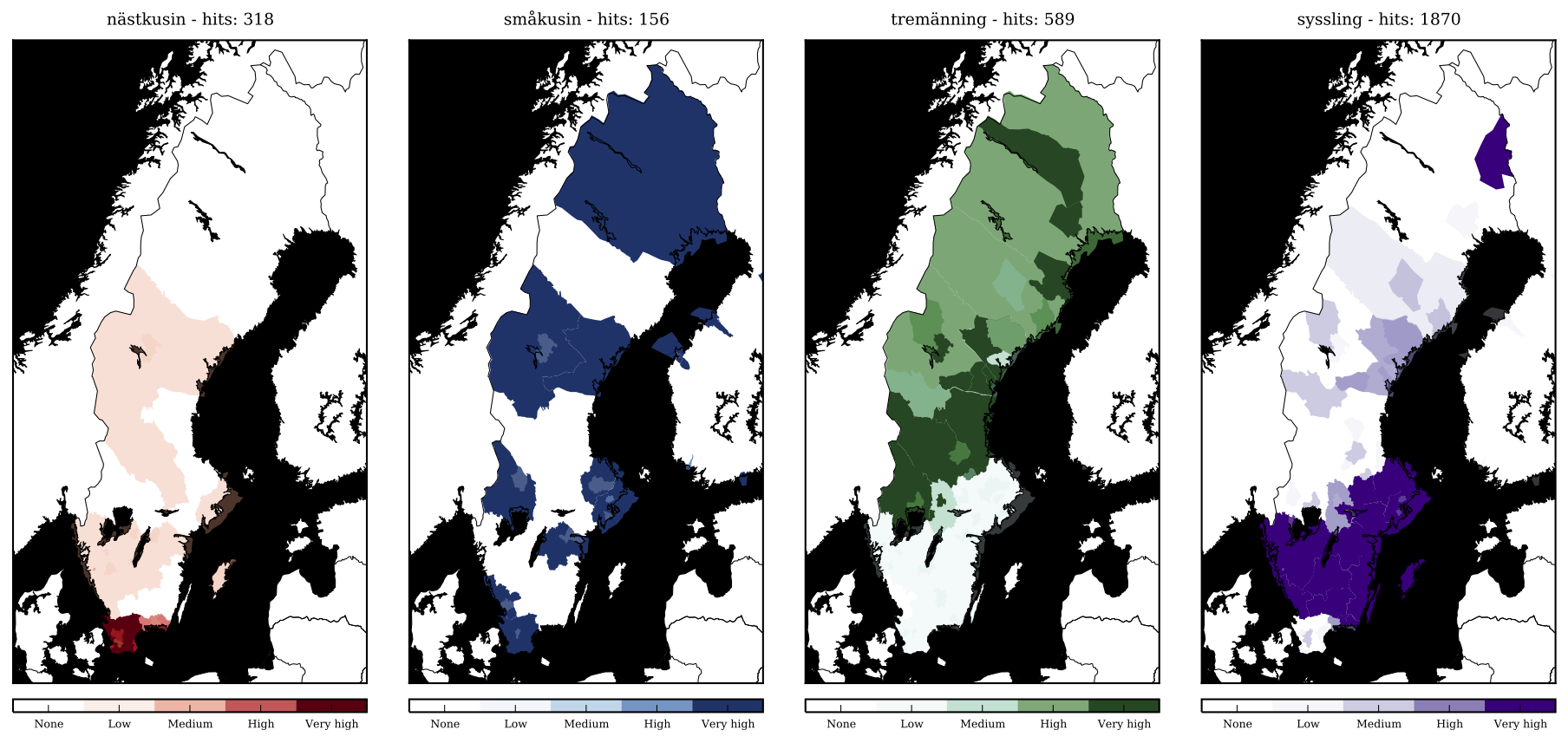}
    \caption{Using labeled data set}

\end{subfigure} 
\begin{subfigure}{0.75\textwidth}
\centering

    \includegraphics[width=\textwidth]{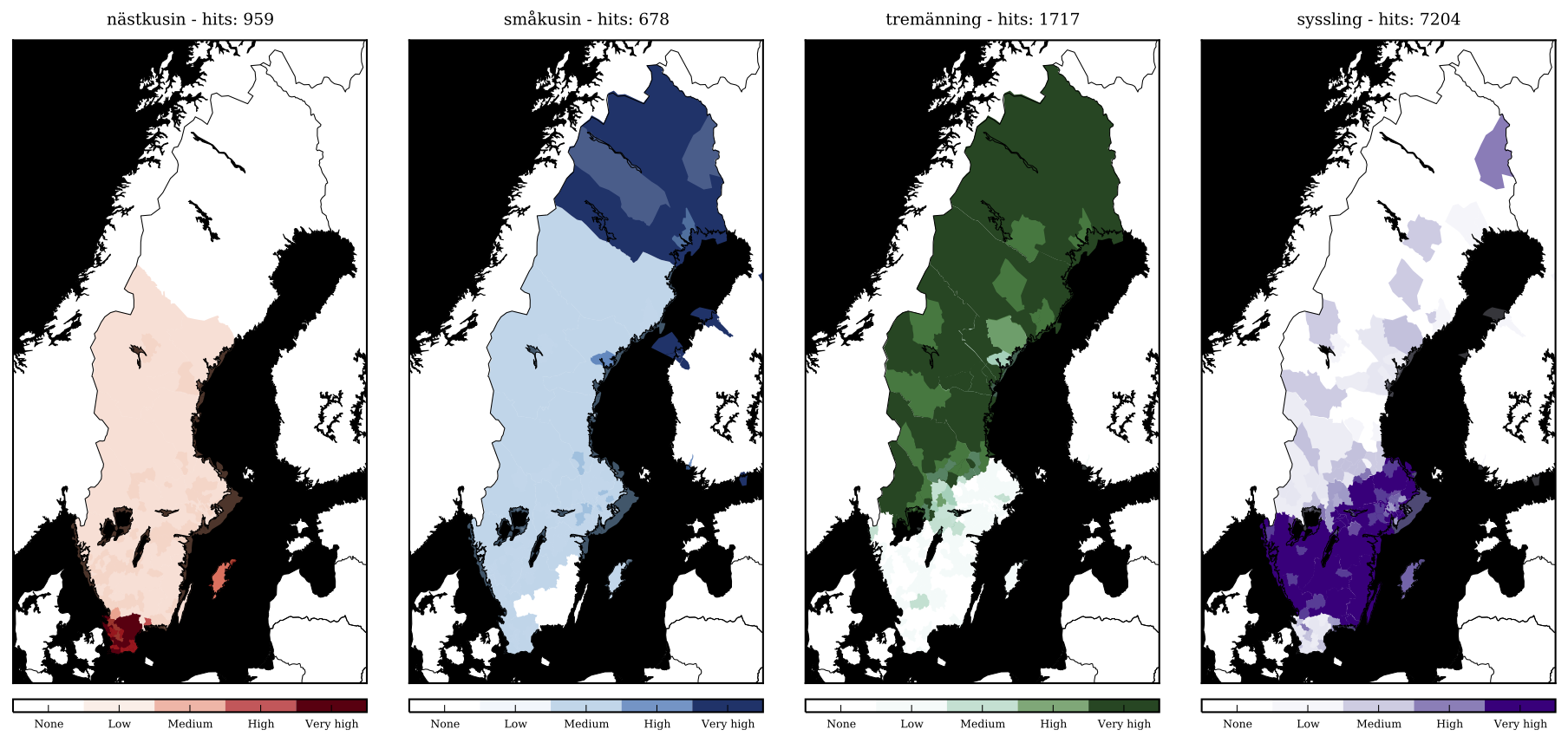}
    \caption{Using enriched data set increases the data}
\end{subfigure}
    \caption{Regional terminology for ``second cousin''}
    \label{cousinmap}
\end{figure*}

\begin{figure*}[p]
\centering
    \includegraphics[width=0.92\textwidth]{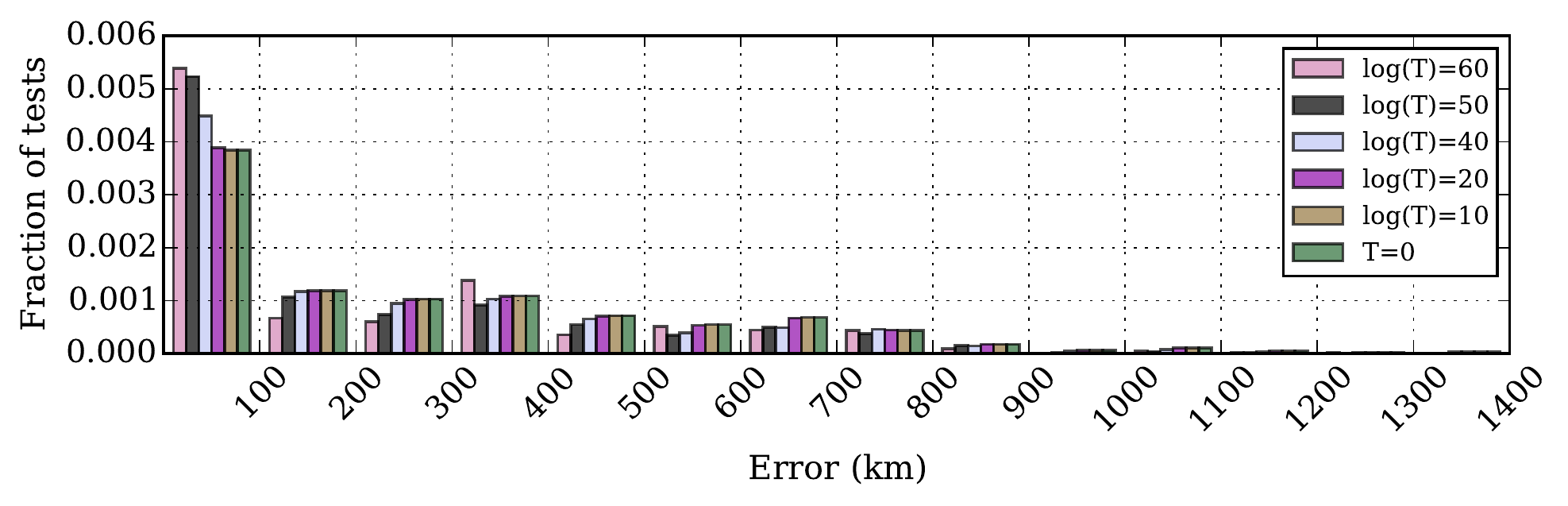}
    \caption{Comparing placeness thresholds for the {\sc Filtered Centroid} model.}
    \label{compthresholds}
\end{figure*}

\begin{table*}[p]
\centering

\makebox[\textwidth][c]{
\begin{tabular}{r|c|cc|ccc|cc}
 & Placeness &\multicolumn{2}{c}{Error (km)} & \multicolumn{3}{c}{Percentile (km)} & \multicolumn{2}{c}{$e < 100~km$} \\
  & $\log T$ &$\tilde{e}$ & $\bar{e}$ & $25~\%$ & $50~\%$ & $75~\%$ & Precision & Recall \\ \hline

{\sc Filtered Centroid} \definecolor{sparkspikecolor}{rgb}{0.18,0.44,0.22}\begin{sparkline}{14}
\sparkspike 0.083 0.987708612414
\sparkspike 0.147153846154 0.305883976632
\sparkspike 0.211307692308 0.264822844128
\sparkspike 0.275461538462 0.279415999234
\sparkspike 0.339615384615 0.184322040799
\sparkspike 0.403769230769 0.140924095876
\sparkspike 0.467923076923 0.175356311518
\sparkspike 0.532076923077 0.113359247342
\sparkspike 0.596230769231 0.0458778274899
\sparkspike 0.660384615385 0.0165961371793
\sparkspike 0.724538461538 0.0272310181878
\sparkspike 0.788692307692 0.011731752144
\sparkspike 0.852846153846 0.00691505715803
\sparkspike 0.917 0.00996722031744
\end{sparkline}
             & --- & 204 & 365 & 45 & 204 & 464 & 0.38 & 0.38  \\         
             
{\sc Filtered Centroid} \definecolor{sparkspikecolor}{rgb}{0.09,0.19,0.29}\begin{sparkline}{14}
\sparkspike 0.083 0.987737974363
\sparkspike 0.147153846154 0.305878300867
\sparkspike 0.211307692308 0.264817930265
\sparkspike 0.275461538462 0.27941081459
\sparkspike 0.339615384615 0.184318620651
\sparkspike 0.403769230769 0.140921480989
\sparkspike 0.467923076923 0.175353057731
\sparkspike 0.532076923077 0.113357143929
\sparkspike 0.596230769231 0.0458769762137
\sparkspike 0.660384615385 0.0165958292332
\sparkspike 0.724538461538 0.0272305129085
\sparkspike 0.788692307692 0.011731534458
\sparkspike 0.852846153846 0.00691492884717
\sparkspike 0.917 0.00996703537282
\end{sparkline}
             & 10 & 204 & 365 & 45 & 204 & 464 & 0.38 & 0.38  \\     
      
      {\sc Filtered Centroid} \definecolor{sparkspikecolor}{rgb}{0.56,0.05,0.67}\begin{sparkline}{14}
\sparkspike 0.083 1.0
\sparkspike 0.147153846154 0.30458313018
\sparkspike 0.211307692308 0.262749878108
\sparkspike 0.275461538462 0.278352023403
\sparkspike 0.339615384615 0.181423695758
\sparkspike 0.403769230769 0.136665041443
\sparkspike 0.467923076923 0.17240370551
\sparkspike 0.532076923077 0.115114578255
\sparkspike 0.596230769231 0.0457825451
\sparkspike 0.660384615385 0.01662603608
\sparkspike 0.724538461538 0.0275962944905
\sparkspike 0.788692307692 0.0117503656753
\sparkspike 0.852846153846 0.0068746952706
\sparkspike 0.917 0.0101901511458
\end{sparkline}
             & 20 & 200 & 365 & 44 & 200 & 460 & 0.38 & 0.38  \\

      {\sc Filtered Centroid} \definecolor{sparkspikecolor}{rgb}{0.76,0.78,0.95}\begin{sparkline}{14}
\sparkspike 0.083 1.15395221717
\sparkspike 0.147153846154 0.300933156223
\sparkspike 0.211307692308 0.243010474372
\sparkspike 0.275461538462 0.265716165657
\sparkspike 0.339615384615 0.168075073394
\sparkspike 0.403769230769 0.100421380992
\sparkspike 0.467923076923 0.12623834776
\sparkspike 0.532076923077 0.119221428587
\sparkspike 0.596230769231 0.0366733322805
\sparkspike 0.660384615385 0.0124451019291
\sparkspike 0.724538461538 0.0199254025568
\sparkspike 0.788692307692 0.00933382644685
\sparkspike 0.852846153846 0.00503099865221
\sparkspike 0.917 0.00913523439479
\end{sparkline}
             & 40 & 145 & 333 & 32 & 145 & 396 & 0.44 & 0.32 \\

      {\sc Filtered Centroid} \definecolor{sparkspikecolor}{rgb}{0.0,0.0,0.0}\begin{sparkline}{14}
\sparkspike 0.083 1.34520175248
\sparkspike 0.147153846154 0.274587355116
\sparkspike 0.211307692308 0.190425905715
\sparkspike 0.275461538462 0.236650944212
\sparkspike 0.339615384615 0.141119197985
\sparkspike 0.403769230769 0.0900059944983
\sparkspike 0.467923076923 0.128048669859
\sparkspike 0.532076923077 0.0964881263334
\sparkspike 0.596230769231 0.0398491711179
\sparkspike 0.660384615385 0.00626960292255
\sparkspike 0.724538461538 0.0104139167188
\sparkspike 0.788692307692 0.00435684270889
\sparkspike 0.852846153846 0.00329419814575
\sparkspike 0.917 0.00340046260206
\end{sparkline}
             & 50 & 90 & 286 & 22 & 90 & 321 & 0.52 & 0.23 \\

      {\sc Filtered Centroid} \definecolor{sparkspikecolor}{rgb}{0.83,0.53,0.71}\begin{sparkline}{14}
\sparkspike 0.083 1.38501580023
\sparkspike 0.147153846154 0.171734093545
\sparkspike 0.211307692308 0.154036305279
\sparkspike 0.275461538462 0.354611238961
\sparkspike 0.339615384615 0.0917663095279
\sparkspike 0.403769230769 0.132405675176
\sparkspike 0.467923076923 0.11405241327
\sparkspike 0.532076923077 0.113396939631
\sparkspike 0.596230769231 0.0242525246609
\sparkspike 0.660384615385 0.00327736819742
\sparkspike 0.724538461538 0.0117985255107
\sparkspike 0.788692307692 0.00524378911588
\sparkspike 0.852846153846 0.00655473639485
\sparkspike 0.917 0.00196642091845
\end{sparkline}
             & 60 & 70 & 271 & 13 & 70 & 330 & 0.53 & 0.04

\end{tabular}
} 

\caption{Comparing placeness thresholds for the {\sc Filtered Centroid} model. }
\label{testspec2}

\end{table*}
\begin{figure*}[p]
\centering
    \includegraphics[width=0.92\textwidth]{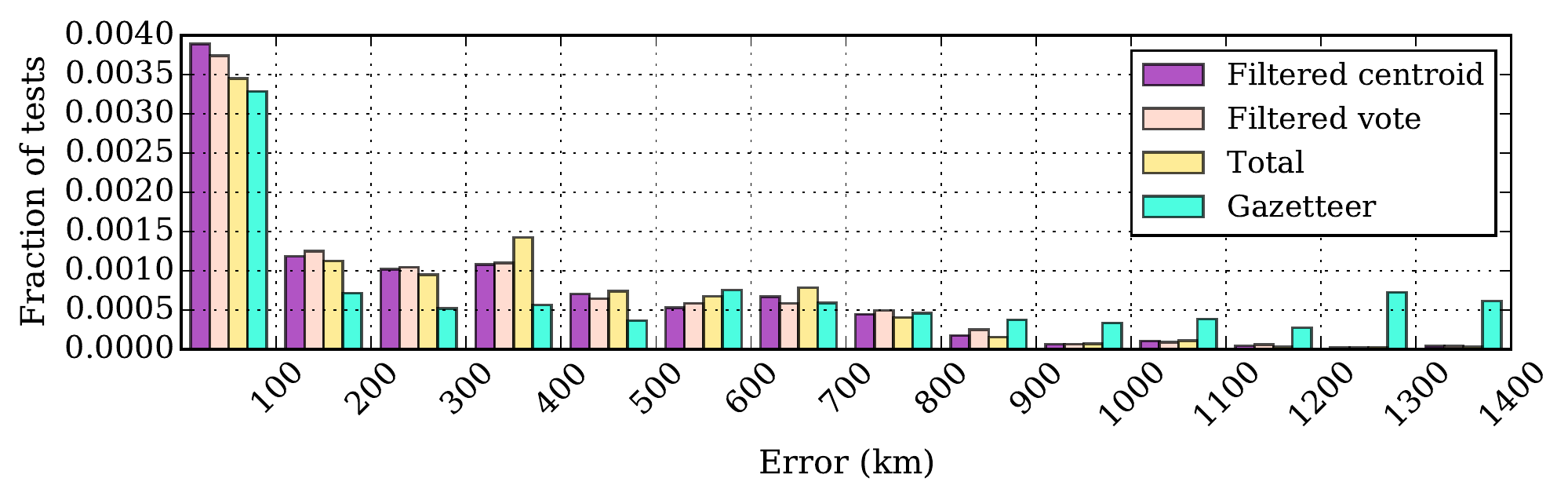}
    \caption{Comparing models with placeness threshold at $\log T=20$.}
    \label{comptests}
\end{figure*}

\begin{table*}[p]
\centering

\makebox[\textwidth][c]{ 
\begin{tabular}{r|c|cc|ccc|cc}
 & Placeness &\multicolumn{2}{c}{Error (km)} & \multicolumn{3}{c}{Percentile (km)} & \multicolumn{2}{c}{$e < 100~km$} \\
  & $\log T$ &$\tilde{e}$ & $\bar{e}$ & $25~\%$ & $50~\%$ & $75~\%$ & Precision & Recall \\ \hline
 
{\sc Gazetteer} \definecolor{sparkspikecolor}{rgb}{0,0.99,0.83} \begin{sparkline}{14}
\sparkspike 0.083 0.845362441969
\sparkspike 0.147153846154 0.184381127077
\sparkspike 0.211307692308 0.134481436256
\sparkspike 0.275461538462 0.145502847197
\sparkspike 0.339615384615 0.0948953593431
\sparkspike 0.403769230769 0.194239728608
\sparkspike 0.467923076923 0.151873020494
\sparkspike 0.532076923077 0.119263800048
\sparkspike 0.596230769231 0.0974232058893
\sparkspike 0.660384615385 0.0868568073263
\sparkspike 0.724538461538 0.100052166297
\sparkspike 0.788692307692 0.0706785894308
\sparkspike 0.852846153846 0.187161758278
\sparkspike 0.917 0.157939852205
\end{sparkline}     
             & 20 & 450 & 626 & 62 & 450 & 964 & 0.31 & 0.31  \\

{\sc Total}
\definecolor{sparkspikecolor}{rgb}{1,0.90,0.42}\begin{sparkline}{14}
\sparkspike 0.083 0.887698346421
\sparkspike 0.147153846154 0.289671993426
\sparkspike 0.211307692308 0.244062237304
\sparkspike 0.275461538462 0.366212503699
\sparkspike 0.339615384615 0.190779366516
\sparkspike 0.403769230769 0.173383795999
\sparkspike 0.467923076923 0.201740958896
\sparkspike 0.532076923077 0.10508830956
\sparkspike 0.596230769231 0.0404625736128
\sparkspike 0.660384615385 0.0184917297547
\sparkspike 0.724538461538 0.0287860947728
\sparkspike 0.788692307692 0.00891225119622
\sparkspike 0.852846153846 0.0071488645959
\sparkspike 0.917 0.00767311466626
\end{sparkline}
             & 20 & 256 & 380 & 51 & 256 & 516 & 0.34 & 0.34  \\
             
{\sc Filtered Centroid} \definecolor{sparkspikecolor}{rgb}{0.56,0.05,0.67}\begin{sparkline}{14}
\sparkspike 0.083 1.0
\sparkspike 0.147153846154 0.30458313018
\sparkspike 0.211307692308 0.262749878108
\sparkspike 0.275461538462 0.278352023403
\sparkspike 0.339615384615 0.181423695758
\sparkspike 0.403769230769 0.136665041443
\sparkspike 0.467923076923 0.17240370551
\sparkspike 0.532076923077 0.115114578255
\sparkspike 0.596230769231 0.0457825451
\sparkspike 0.660384615385 0.01662603608
\sparkspike 0.724538461538 0.0275962944905
\sparkspike 0.788692307692 0.0117503656753
\sparkspike 0.852846153846 0.0068746952706
\sparkspike 0.917 0.0101901511458
\end{sparkline}
             & 20 & 200 & 365 & 44 & 200 & 460 & 0.38 & 0.38  \\

{\sc Filtered Vote} \definecolor{sparkspikecolor}{rgb}{1,0.78,0.71}\begin{sparkline}{14}
\sparkspike 0.083 0.961332997736
\sparkspike 0.147153846154 0.321639538255
\sparkspike 0.211307692308 0.269599637067
\sparkspike 0.275461538462 0.2832637452
\sparkspike 0.339615384615 0.166634283227
\sparkspike 0.403769230769 0.151177366579
\sparkspike 0.467923076923 0.150353643749
\sparkspike 0.532076923077 0.12719249592
\sparkspike 0.596230769231 0.0648318322062
\sparkspike 0.660384615385 0.0176373594343
\sparkspike 0.724538461538 0.0240333249434
\sparkspike 0.788692307692 0.0155538252154
\sparkspike 0.852846153846 0.00658978264577
\sparkspike 0.917 0.0102723082419
\end{sparkline}      
             & 20 & 208 & 377 & 58 & 208 & 467 & 0.37 & 0.36

\end{tabular}
} 

\caption{Comparing models: $\tilde{e}$ is the median error and $\bar{e}$ is the mean error in km.}
\label{testspec}

\end{table*}


\section{Experimental settings: the {\sc Total} and {\sc Filtered} models}

We run one experimental setting with all words of a set, only filtered for placeness. We call this approach the {\sc Total} approach.

As a more informed model, we filter the words in the feature set to find the most locationally appropriate terms, in order to reduce noise and computational effort, but above all, in keeping with our hypothesis that the locational signal is present in only part of the texts. \citet{backstrom2008spatial} and following them, \citet{cheng2010you}, using similar data as we do, also limit their analyses to ``local'' rather than ``non-local'' words in the text matter they process, modeling word locality through observed occurrences, modulated with some geographical smoothing. To find the most appropriate localised linguistic items, we bootstrap from the gazetteer and collect the most distinctive distributional contexts of gazetteer terms. For this, we used context windows of six words before ($6+0$), around ($3+3$), and after ($0+6$) each target word. These context windows were tabulated and the most frequently occurring constructions\footnote{In these experiments, the 900 most frequent constructions are used.} are then ranked based on their ability to return words with high placeness. For each construction, the percentage of words returned with $\log T > 20$ is used as a ranking criterion. Using this ranking, the top 150 constructions are retained as a paradigmatic filter to generate usefully locational words. Constructions such as {\tt lives in <location>} will be at the top of the list. Examples are given in Figure~\ref{patternexamples}.

\begin{figure}[h]
\centering
\scriptsize
\begin{subfigure}{0.23\textwidth}
    {\tt <location> mellan \\
varit i <location> \\
bor i <location> \\
var i <location> \\
vi till <location> \\
in till <location> \\
ska till <location> \\
<location> centrum \\
av till <location> \\
det av till <location> \\
hemma i <location> \\
till <location> \\
upp till <location> }
\caption{In Swedish}
\end{subfigure} \quad \begin{subfigure}{0.20\textwidth}
    {\tt <location> between \\
been in <location> \\
live(s) in <location> \\
was in <location> \\
we to <location> \\
in to <location> \\
going to <location> \\
<location> centre \\
off to <location> \\
go to <location> \\
home in <location> \\
to <location> \\
up to <location> }
\caption{Translated to English}
\end{subfigure}
\caption{Examples of locational constructions} \label{patternexamples}
\end{figure}

Words found in the {\tt <location>} slot of the constructions are frequency filtered with respect to $N$, the length of the text under analysis, with thresholds set by experimentation to $0.00008 \times N \leq f_{wd} \leq N / 300$. This reduces the number of Gaussian models to evaluate drastically. Each text under consideration was then filtered to only include words found through the above procedure, reducing the size of the texts to about 6\% of the original.

\section{Aggregating the locational information for filtered texts}

The filtered texts are now processed in two different ways. Every unique word token in the Twitter dataset has a Gaussian mixture model $i$ based on its observed occurrences, as shown in Section~\ref{placeness}. This is represented by the three mean coordinates $\overline{\mu}^i$ and their corresponding {\it placenesses} $\overline{p}^i$.

\begin{align*}
\overline{\mu}^i & = \begin{pmatrix} \mu_1 \\  \mu_2 \\  \mu_3 \end{pmatrix}^i & \overline{p}^i & = \begin{pmatrix} p_1 \\  p_2 \\  p_3 \end{pmatrix}^i  
\end{align*}

We compute a centroid for these coordinates, as an average best guess for geographic signal for a text. We do this with an arithmetic weighted mean. Given $n$ words:

$$
M = \frac{ \sum\limits_{i=1}^n \overline{\mu}^n \cdot \overline{p}^n }{\sum\limits_{i=1}^n \sum\limits_{j=1}^3 p^n_j}
$$ 

Where $\overline{\mu}^n \cdot \overline{p}^n$ is the dot product\footnote{$\overline{\mu}^i \cdot \overline{p}^i = \mu_1^i p_1^i + \mu_2^i p_2^i + \mu_3^i p_3^i$  for this specific case.}. We call this model {\sc Filtered centroid}

Alternatively, we do not average the coordinates, but select by weighted majority vote. We divide Sweden into a grid of roughly 50x50km cells. The placeness score of every locational word in a text is added to its cell. The centerpoint of the cell with highest score is assigned to the text as a location. We call this model {\sc Filtered Vote}.

Figure~\ref{gridexample} shows how filtering improves results, here illustrated by the {\sc Filtered Vote} model. The top map shows how every word of a text contributes votes, weighted by their placeness, to give a prediction (\includegraphics[width=0.015\textwidth]{reddot.png}). The bottom map shows how when only words filtered through the distributional model are used, the voting yields a correct result in comparison with the gold standard (\includegraphics[width=0.015\textwidth]{greendot.png}) given by the metadata.

\section{Results}

As shown in Table~\ref{testspec} and Figure~\ref{comptests}, the Gaussian models {\sc Filtered Centroid}\mbox{\definecolor{sparkspikecolor}{rgb}{0.56,0.05,0.67}\begin{sparkline}{14}
\sparkspike 0.083 1.0
\sparkspike 0.147153846154 0.30458313018
\sparkspike 0.211307692308 0.262749878108
\sparkspike 0.275461538462 0.278352023403
\sparkspike 0.339615384615 0.181423695758
\sparkspike 0.403769230769 0.136665041443
\sparkspike 0.467923076923 0.17240370551
\sparkspike 0.532076923077 0.115114578255
\sparkspike 0.596230769231 0.0457825451
\sparkspike 0.660384615385 0.01662603608
\sparkspike 0.724538461538 0.0275962944905
\sparkspike 0.788692307692 0.0117503656753
\sparkspike 0.852846153846 0.0068746952706
\sparkspike 0.917 0.0101901511458
\end{sparkline}}and {\sc Filtered Vote}\mbox{\definecolor{sparkspikecolor}{rgb}{1,0.78,0.71}\begin{sparkline}{14}
\sparkspike 0.083 0.961332997736
\sparkspike 0.147153846154 0.321639538255
\sparkspike 0.211307692308 0.269599637067
\sparkspike 0.275461538462 0.2832637452
\sparkspike 0.339615384615 0.166634283227
\sparkspike 0.403769230769 0.151177366579
\sparkspike 0.467923076923 0.150353643749
\sparkspike 0.532076923077 0.12719249592
\sparkspike 0.596230769231 0.0648318322062
\sparkspike 0.660384615385 0.0176373594343
\sparkspike 0.724538461538 0.0240333249434
\sparkspike 0.788692307692 0.0155538252154
\sparkspike 0.852846153846 0.00658978264577
\sparkspike 0.917 0.0102723082419
\end{sparkline}}outperform the {\sc Gazetteer} model\mbox{ \definecolor{sparkspikecolor}{rgb}{0,0.99,0.83}\begin{sparkline}{14}
\sparkspike 0.083 0.845362441969
\sparkspike 0.147153846154 0.184381127077
\sparkspike 0.211307692308 0.134481436256
\sparkspike 0.275461538462 0.145502847197
\sparkspike 0.339615384615 0.0948953593431
\sparkspike 0.403769230769 0.194239728608
\sparkspike 0.467923076923 0.151873020494
\sparkspike 0.532076923077 0.119263800048
\sparkspike 0.596230769231 0.0974232058893
\sparkspike 0.660384615385 0.0868568073263
\sparkspike 0.724538461538 0.100052166297
\sparkspike 0.788692307692 0.0706785894308
\sparkspike 0.852846153846 0.187161758278
\sparkspike 0.917 0.157939852205
\end{sparkline}}handily. Filtering words distributionally, in addition to reducing processing, improves results further. The {\sc Filtered Centroid} model\mbox{\definecolor{sparkspikecolor}{rgb}{0.56,0.05,0.67}\begin{sparkline}{14}
\sparkspike 0.083 1.0
\sparkspike 0.147153846154 0.30458313018
\sparkspike 0.211307692308 0.262749878108
\sparkspike 0.275461538462 0.278352023403
\sparkspike 0.339615384615 0.181423695758
\sparkspike 0.403769230769 0.136665041443
\sparkspike 0.467923076923 0.17240370551
\sparkspike 0.532076923077 0.115114578255
\sparkspike 0.596230769231 0.0457825451
\sparkspike 0.660384615385 0.01662603608
\sparkspike 0.724538461538 0.0275962944905
\sparkspike 0.788692307692 0.0117503656753
\sparkspike 0.852846153846 0.0068746952706
\sparkspike 0.917 0.0101901511458
\end{sparkline}}is slightly better than the {\sc Filtered Vote} model\mbox{\definecolor{sparkspikecolor}{rgb}{1,0.78,0.71}\begin{sparkline}{14}
\sparkspike 0.083 0.961332997736
\sparkspike 0.147153846154 0.321639538255
\sparkspike 0.211307692308 0.269599637067
\sparkspike 0.275461538462 0.2832637452
\sparkspike 0.339615384615 0.166634283227
\sparkspike 0.403769230769 0.151177366579
\sparkspike 0.467923076923 0.150353643749
\sparkspike 0.532076923077 0.12719249592
\sparkspike 0.596230769231 0.0648318322062
\sparkspike 0.660384615385 0.0176373594343
\sparkspike 0.724538461538 0.0240333249434
\sparkspike 0.788692307692 0.0155538252154
\sparkspike 0.852846153846 0.00658978264577
\sparkspike 0.917 0.0102723082419
\end{sparkline}}, providing support for late discretization of locational information. A closer look at the effect, shown in Table~\ref{testspec2} and in Figure~\ref{compthresholds}, of feature selection with the placeness threshold shows the precision-recall tradeoff contingent on reducing the number of accepted locational words. 

These results are well comparable with the results reported by others: while direct comparison with other linguistic and geographic areas is difficult, \citet{cheng2010you} set a 100-mile ($\approx$ 160 km) success criterion for a similar task of geo-locating microblog authors (not single posts). They find that about 10\% of microblog users can be localised within their 100-mile radius. \citet{eisenstein2010latent} found they could on average achieve a 900 km accuracy for texts or a 24\% accuracy on a US state level. 

\section{Regional variation}

Returning to our use case we now use the {\sc Filtered Centroid} model\mbox{\definecolor{sparkspikecolor}{rgb}{0.56,0.05,0.67}\begin{sparkline}{14}
\sparkspike 0.083 1.0
\sparkspike 0.147153846154 0.30458313018
\sparkspike 0.211307692308 0.262749878108
\sparkspike 0.275461538462 0.278352023403
\sparkspike 0.339615384615 0.181423695758
\sparkspike 0.403769230769 0.136665041443
\sparkspike 0.467923076923 0.17240370551
\sparkspike 0.532076923077 0.115114578255
\sparkspike 0.596230769231 0.0457825451
\sparkspike 0.660384615385 0.01662603608
\sparkspike 0.724538461538 0.0275962944905
\sparkspike 0.788692307692 0.0117503656753
\sparkspike 0.852846153846 0.0068746952706
\sparkspike 0.917 0.0101901511458
\end{sparkline}}to position and thus enrich a further 38\% of our unlabeled blog collection with a location tag (setting the placeness threshold $\log T = 20$). This gives a noticeably better resolution for studying regional word usage as shown in Figure~\ref{cousinmap}: the term for ``second cousin'' varies across dialects, and given the enriched data set we are able to gain better frequencies and a more distinct image of usage.

\section{Conclusions}

We find that
\begin{itemize}
\item modelling geographical distribution of linguistic items with multiple (in this case, three) peaks proved useful;
\item filtering locationally indicative linguistic items using distributional constructions proved useful;
\item modelling the placeness of locational linguistic items for thresholding proved useful; 
\item training a locational model on positionally annotated microblog posts was a useful bootstrap for assigning location to texts of an entirely different genre;
\item we are able to detect and explore regional variation in terminological usage.

\end{itemize}

\section*{Acknowlegdments} This work was in part supported by the grant SINUS (Spridning av innovationer i nutida svenska) from Vetenskapsr{\aa}det, the Swedish Research Council.

\bibliographystyle{plainnat}

\end{document}